\documentclass[preprint,12pt]{elsarticle}

\usepackage{amssymb}

\begin{document}

\begin{frontmatter}



\title{{\small Comment} \\ \vspace{0.5cm} Beyond description. Comment on "Approaching human language with complex networks" by Cong \&  Liu}


\author{R. Ferrer-i-Cancho}

\ead{rferrericancho@cs.upc.edu}

\address{Complexity \& Quantitative Linguistics Lab, LARCA Research Group, Departament de Ci\`encies de la Computaci\'o, Universitat Polit\`ecnica de Catalunya, Campus Nord, Edifici Omega Jordi Girona Salgado 1-3. 08034 Barcelona, Catalonia (Spain)}







\end{frontmatter}



In their historical overview, Cong \& Liu highlight Sausurre as the father of modern linguistics \cite{Cong2014a}. They apparently miss
G. K. Zipf as a pioneer of the view of language as a complex system. His idea of a balance between unification and diversification forces in the organization of natural systems, e.g., vocabularies \cite{Zipf1949a}, can be seen as a precursor of the view of complexity as a balance between order (unification) and disorder (diversification) near the edge of chaos \cite{Kauffman1993}. Although not mentioned by Cong \& Liu somewhere else, trade-offs between hearer and speaker needs are very important in Zipf's view, which has inspired research on the optimal networks mapping words into meanings \cite{Ferrer2007a,Prokopenko2010a,Ferrer2013g}. 
Quantitative linguists regard G. K. Zipf as the funder of modern quantitative linguistics \cite{Altmann2002a}, a discipline where statistics plays a central role as in network science. Interestingly, that centrality of statistics is missing Saussure's work and that of many of his successors.

Cong \& Liu review the relationship between the sentence and the global level in syntactic dependency networks \cite{Ferrer2003f} through randomizations of the syntactic structures of the sentences that are used to build a global syntactic dependency network \cite{Liu2008b}. Cong \& Liu argue that the randomization procedure produces non-syntactic global networks, i.e. they take for granted that all what the randomization procedure preserves is purely non-syntactic (notice that the randomization method preserves word frequencies). 
However, the strong positive correlation between word frequency and word degree, even when sentence structures are randomized, predicts a distribution of degrees that mirrors Zipf's law \cite{Ferrer2003f,Masucci2009a}. That degree distribution includes hubs reminiscent of function words \cite{Ferrer2004f}. To destroy syntax completely, one has to remove not only those protosyntactic hubs but also suppress any syntactic factor from the word frequencies that the original randomization procedure preserves (e.g., the need of combining content words or expressing relationships). From a generativist view, word frequencies are anecdotal \cite{Miller1963}. From a non-reductionistic view taking into account grammaticalization processes, word frequency and syntax are harder to separate \cite{Ferrer2004f}.
     
Cong \& Liu have shown convincingly that network science provides many new perspectives for investigating language as a complex system. 
They are concerned about the statistical properties that have been reported usually for linguistic networks (heterogeneous degree distribution, small-worldness...) but pay little attention to general requirements of a complex enough system. 
A complex system needs to join a large proportion of elements of the system. 
A network where only a few vertices are connected cannot be called complex, as it cannot be said that a child producing 1-word or 2-word utterances exhibits a complex language \cite{Saxton2010a}. Such a network is not a small world: the majority of vertices are at infinite distance. Real complex networks are not that way. 
Consider the other extreme, a network where all vertices are connected (a complete graph) would show what many authors consider to be the small-world property: high clustering (the clustering would be one, the maximum that is possible) and small world (all vertices would be at distance one, the minimum distance possible that is possible). However, a network that is that dense may not be very functional (vertices would be connecting too freely and lack distinguishable roles) and it might be too expensive to build (imagine that links have some cost and that the network has many vertices: the cost would be maximum). Not surprisingly, children never end up connecting every word with every other word. This raises the question of whether the linguistic networks of a child cross a percolation phase transition as they develop \cite{Ferrer2003_Thesis_Introduction}. 
A complex system needs that its network has a large enough connected component. Interestingly, connectedness has been considered a requirement for a rudimentary form of syntax and symbolic reference \cite{Ferrer2004f}. As real linguistic (and non-linguistic) networks are sparse (as noted by Cong \& Liu), connectedness or the presence of the giant component are important aspects from a systems theoretic perspective. Those features appear as candidates for more fundamental universals of fully-fledged languages than typical measures from popular network science. 

Cong \& Liu's review presents network theory essentially as new perspectives for linguistic description \cite{Cong2014a} but network theory goes beyond that providing components for a new theoretical linguistics: models of network dynamics suggesting principles for the formation of linguistic networks and ground-breaking discoveries about the dynamics on networks of social interactions \cite{Baronchelli2013a}. As for the later, Loreto and collaborators stroke the internalist bias of language research (not only of theoretical linguistics) showing that the emergence of a shared communication system within a population depends on the network properties of the interactions among individuals \cite{Dallasta2006a}.
Network theory not only brings us new theoretical questions but suggests answers to old and recurring questions about structural properties of language: if one-to-one mappings between words and meanings are convenient for communication why are real word-meaning mappings not that way? \cite{Ferrer2002a} Why do children prefer to attach new words to unlinked meanings? \cite{Ferrer2013g} Why do syntactic links tend to not cross when drawn over the sentence? \cite{Ferrer2006d}. Network theory not only defines binomial networks as baselines for average shortest path lengths but also random and minimum linear arrangements of vertices for investigating the limits of the variation of mean dependency length and the number of crossings in syntactic dependency trees \cite{Ferrer2004b,Ferrer2013d}. Baselines of this kind are vital for drawing solid conclusions about optimization in language \cite{Ferrer2013c,Ferrer2013d} and are just one example of the new mathematical frameworks for linguistic theory that have been developed from network theory.

Having said that, Cong \& Liu have made a tremendous effort to organize and classify the vast literature of linguistic networks research, illustrated beautifully how networks can be used for typological research (language classification), reviewed the new understanding of function words from network theory and so on. Cong \& Liu have not hesitated to comment on the limits of current research, from methodological issues to a lack of linguistic depth. For instance, they call the attention of the reader on the need of comparing languages with parallel corpora to control better for linguistic variables such as genre, topic and so on. However, they omit more elementary and widely applicable controls such as dataset size. The point is that certain network theoretic metrics may scale with the size of the source dataset (and thus one may mistake differences in dataset size for real differences between languages). And I totally agree with Ninio's concerns about the problems of cooccurrence networks \cite{Ninio2014a}. 
 
Cong \& Liu are totally fair when noting the need of introducing more linguistics in the motivation and also in the interpretation of results of network research. They call for more linguistic detail in analyses, e.g., comparing the impact of having word forms or lemmas as vertices. 
When reviewing thoroughly studies of Chinese networks, Cong \& Liu seem to invite us to get rid of the bias for Indo-European languages (e.g., English) in theoretical and empirical research. Cong \& Liu are defining the linguistics of the future based on a network approach to language that feeds, renews and unifies theoretical linguistics. 




\section*{Acknowledgments}
This work was supported by the grant BASMATI (TIN2011-27479-C04-03) from the Spanish Ministry of Science and Innovation.


\bibliographystyle{elsarticle-num} 

\end{document}